# Join-graph based cost-shifting schemes

**Alexander Ihler, Natalia Flerova, Rina Dechter and Lars Otten**
University of California Irvine
[ihler, nflerova, dechter, lotten]@ics.uci.edu

## Abstract

We develop several algorithms taking advantage of two common approaches for bounding MPE queries in graphical models: mini-bucket elimination and message-passing updates for linear programming relaxations. Both methods are quite similar, and offer useful perspectives for the other; our hybrid approaches attempt to balance the advantages of each. We demonstrate the power of our hybrid algorithms through extensive empirical evaluation. Most notably, a Branch and Bound search guided by the heuristic function calculated by one of our new algorithms has recently won first place in the PASCAL2 inference challenge.

## 1 INTRODUCTION

Combinatorial optimization tasks such as finding the most likely variable assignment of a probability model, the *most probable explanation* (MPE) or maximum *a posteriori* (MAP) problem, arise in many applications. These problems are typically NP-hard; graphical models provide a popular framework for reasoning about such tasks and organizing computations (Pearl, 1988).

Mini-Bucket Elimination (MBE) (Dechter & Rish, 2003) is a popular bounding scheme that generates upper and lower bounds by applying the exact Bucket Elimination (BE) algorithm (Dechter, 1999) to a simplified problem obtained by duplicating variables. The relaxation view of MBE is closely related to a family of iterative approximation techniques based on linear programming (LP). These include "reweighted" max-product (Wainwright *et al.*, 2005), max-product linear programming (MPLP) (Globerson & Jaakkola, 2007), dual decomposition (Komodakis *et al.*, 2007), and soft arc consistency (Schiex, 2000; Bistarelli *et al.*, 2000). These algorithms simplify the graphical model into independent components and tighten the resulting bound via iterative cost-shifting updates. They can be thought of as "re-parameterizing" the original functions without changing the global model. Most of the schemes operate on the original factors of the model, although some works tighten the approximations by introducing larger clusters (Sontag *et al.*, 2008).

In this paper we combine these ideas to define two new hybrid schemes. One algorithm, called mini-buckets with max-marginal matching (MBE-MM), is a non-iterative algorithm that applies a single pass of cost-shifting during the mini-bucket construction. The second approach, Join Graph Linear Programming (JGLP) applies cost-shifting updates to the full mini-bucket join-graph, iteratively. Our empirical evaluation demonstrates the increased power of these hybrid approximations over their individual components.

Finally, one of the primary uses of bounding schemes is in generating heuristics for Best-First and Branch and Bound search (Marinescu & Dechter, 2009a,b). Our hybrid schemes drastically decrease the search space explored by Branch and Bound to significantly increase its power, evidenced by both our empirical evaluation and the results of the current Probabilistic Inference Challenge[1], where our algorithm won first place in all optimization categories.

From an LP perspective, JGLP can be viewed as a specific algorithm within the class of generalized EMPLP (Sontag & Jaakkola, 2009). Its main novelty is in showing how the systematic join-graph structures of mini-bucket can facilitate effective clique choices. JGLP works "top down", creating large clusters immediately, compared to other methods (e.g., Sontag & Jaakkola, 2009; Batra *et al.*, 2011) that work "bottom up" (gradually including triplets, etc.) and may be less effective for large-width problems. Our non-iterative MBE-MM algorithm is an MBE scheme with some cost-shifting done locally in each bucket. From this perspective it is closely related to (Rollon & Larrosa, 2006); the two methods differ primarily in the form of the cost-shifting update within each bucket. However, our update is motivated by its connection to a globally applicable tightening algorithm.

---

[1]http://www.cs.huji.ac.il/project/PASCAL/

## 2 PRELIMINARIES

We consider combinatorial optimization problems expressed as graphical models, including Markov and Bayesian networks (Pearl, 1988) and constraint networks (Dechter, 2003). A *graphical model* is a tuple $\mathcal{M} = (\mathbf{X}, \mathbf{D}, \mathbf{F}, \bigotimes)$, where $\mathbf{X} = \{X_i : i \in V\}$ is a set of variables indexed by set $V$ and $\mathbf{D} = \{D_i : i \in V\}$ is the set of their finite domains of values. $\mathbf{F} = \{f_\alpha : \alpha \in F\}$ is a set of discrete functions, where we use $\alpha \subseteq V$ and $X_\alpha \subseteq \mathbf{X}$ to indicate the scope of function $f_\alpha$, i.e., $X_\alpha = \text{var}(f_\alpha) = \{X_i : i \in \alpha\}$. The set of function scopes implies a *primal graph* whose vertices are the variables and which includes an edge connecting any two variables that appear in the scope of the same function. The combination operator $\bigotimes \in \{\prod, \sum, \bowtie\}$ defines the complete function represented by the graphical model $\mathcal{M}$ as $C(\mathbf{X}) = \bigotimes_{\alpha \in F} f_\alpha(X_\alpha)$. In this work, we focus on *max-sum problems*, in which we would like to compute the optimal value $C^*$ and/or its optimizing configuration $x^*$:

$$C^* = C(x^*) = \max_{\mathbf{X}} \sum_{\alpha \in F} f_\alpha(X_\alpha) \qquad (1)$$

### 2.1 MINI-BUCKET ELIMINATION

**Bucket elimination** (BE) (Dechter, 1999) is a popular algorithm for solving reasoning tasks over graphical models. It is a special case of cluster tree elimination in which the tree-structure upon which messages are passed is determined by the variable elimination order used. In BE terminology, the nodes of the tree-structure are referred to as buckets and each bucket is associated with a variable to be eliminated.

Each bucket is processed by BE in two steps. First, all functions in the bucket are combined (by summation in the case of max-sum problem). Then the variable associated with the bucket is eliminated from the combined function (by maximization in case of max-sum task). The function resulting from the combination and elimination steps can be viewed as a "message" $\lambda_i$ and is passed to the parent of the current bucket. Processing occurs in this fashion, from the leaves of the tree to the root, one node (bucket) at a time. The time and space complexity of BE are exponential in the graph parameter called the induced width $w$ along the ordering $o$ (Dechter, 1999).

**Mini-bucket elimination** (MBE) (Dechter & Rish, 2003) is an approximation scheme designed to avoid the space and time complexity of BE. Consider a bucket $\mathbf{B}_i$ and an integer bounding parameter $z$. MBE creates a $z$-partition $Q_i = \{q_i^1, \ldots, q_i^p\}$ of $\mathbf{B}_i$, where each set of functions $q_i^j \in Q_i$, called a *mini-bucket*, includes no more than $z + 1$ variables. Then each mini-bucket is processed separately, just as in BE. The scheme generates an upper bound on the exact optimal solution and its time and space complexity of MBE is exponential in $z$, which is chosen to be less than $w$, when $w$ is too large. In general, increasing $z$ tightens the upper bound, until $z = w$, in which case MBE finds the exact solution. The form of mini-bucket makes it easy to estimate and control both the computational and storage complexity through a single parameter $z$.

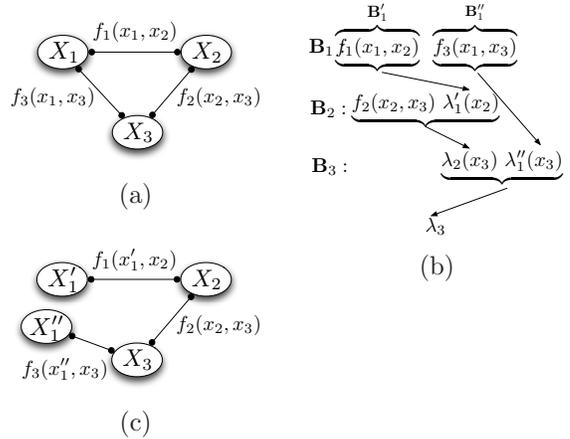

Figure 1: The mini-bucket procedure for a simple graph. (a) Original graph; (b) the buckets and messages computed in MBE; (c) interpreting MBE as variable duplication, and message passing on the resulting junction tree. $X_1$ is duplicated in each of two mini-buckets $q_1^1 = \{f_1(X_1, X_2)\}$ and $q_1^2 = \{f_3(X_1, X_3)\}$.

MBE's value is not only as a stand-alone bounding scheme, but also as a mechanism for generating powerful heuristic evaluation functions for informed search algorithms. Exploration of its potential as a heuristic generation scheme has yielded some of the most powerful Best-First and Branch and Bound search engines for graphical models, as summarized in (Kask & Dechter, 2001; Marinescu & Dechter, 2009a,b).

Mini-bucket elimination can also be interpreted using a junction tree view. The MBE bound corresponds to a problem relaxation in which a copy $X_i^j$ of variable $X_i$ is made for each mini-bucket $q_i^j \in Q_i$, and the resulting messages $\lambda_i^j$ correspond to the messages in a junction tree defined on the augmented model over the variable copies; this junction tree is guaranteed to have induced-width $z$ or less. Figure 1 shows a simple example on three variables. The problems are equivalent if all copies of $X_i$ are constrained to be equal;

**Algorithm 1** Join-Graph Structuring(z)

**Input:** Graphical model $\langle \mathbf{X}, \mathbf{D}, \mathbf{F}, \sum \rangle$, parameter $z$
**Output:** Join-graph with cluster size $\leq z+1$
 1: Order the variables from $X_1$ to $X_n$ minimizing (heuristically) induced-width
 2: Generate an ordered partition of functions $\mathbf{F} = \{f_\alpha\}$ into buckets $\mathbf{B}_1, \ldots, \mathbf{B}_n$, where $\mathbf{B}_i$ is a bucket of variable $X_i$
 3: **for** $i \leftarrow n$ down to 1 (Processing bucket $\mathbf{B}_i$) **do**
 4:  Partition functions in $\mathbf{B}_i$ into $Q_i = \{q_i^1, \ldots, q_i^p\}$; each $q_i^k$ has no more than $z+1$ variables.
 5:  For each mini-bucket $q_i^k$ create a new set of variables $S_i^k = \{X | X \in q_i^k\} - \{X_i\}$ and place it in the bucket of its highest variable in the ordering
 6:  Maintain an arc between $q_i^k$ and the mini-bucket (created later) that includes $S_i^k$
 7: **end for**
 8: Associate each resulting mini-bucket with a node in the join-graph
 9: **Creating arcs:** keep the arcs created in step 6 and also connect the mini-bucket clusters belonging to the same bucket (for example, in a chain).

---

otherwise, the additional degrees of freedom lead to a relaxed problem and thus an upper bound.

The mini-bucket tree has several desirable properties (Mateescu *et al.*, 2010), including that it is a join-graph (satisfies the running intersection property) and each cluster has at most $z+1$ variables. The join-graph construction process is given in Algorithm 1.

### 2.2 LINEAR PROGRAMMING METHODS

Recently various iterative re-parameterization approaches have been introduced that are derived by solving an Linear Programming (LP) relaxation of the graphical model. Wainwright et al. (Wainwright *et al.*, 2005) established the connections between LP relaxations of integer programming problems and (approximate) dynamic programming methods using message-passing in the max-product algebra; subsequent improvements in algorithms such as max-product linear programming (MPLP) include coordinate-descent updates that ensure convergence (Globerson & Jaakkola, 2007; Sontag & Jaakkola, 2009). These methods are closely related to the mini-bucket bounds; since we build upon these approaches we introduce some of the ideas in more detail here.

To match the bulk of the graphical model LP relaxation literature, for this section we assume that the network consists of only pairwise functions $f_{ij}(X_i, X_j)$. A simple bound on the max-sum objective is then given by maxima of the individual functions:

$$C^* = \max_{\mathbf{X}} \sum_{(ij) \in F} f_{ij}(X_i, X_j) \leq \sum_{(ij) \in F} \max_{\mathbf{X}} f_{ij}(X_i, X_j), \quad (2)$$

exchanging the sum and max operators. One can interpret this operation as making an individual copy of each variable for each function, and optimizing over them separately.

However, we can also introduce a collection of functions $\{\lambda_{ij}(X_i), \lambda_{ji}(X_j)\}$ for each edge $(ij)$, and require

$$\lambda \in \Lambda \quad \Leftrightarrow \quad \forall i, \quad \sum_j \lambda_{ij}(X_i) = 0$$

Then, we have

$$\begin{aligned} C^* &= \max_{\mathbf{X}} \sum_{(ij) \in F} f_{ij}(X_i, X_j) \\ &= \max_{\mathbf{X}} \sum_{(ij) \in F} f_{ij}(X_i, X_j) + \sum_i \sum_j \lambda_{ij}(X_i) \\ &\leq \min_{\lambda \in \Lambda} \sum_{(ij) \in F} \max_{\mathbf{X}} \left( f_{ij}(X_i, X_j) + \lambda_{ij}(X_i) + \lambda_{ji}(X_j) \right) \end{aligned} \quad (3)$$

by distributing each $\lambda_{ij}$ to its associated factor and applying the inequality (2).

The new functions $\tilde{f}_{ij} = f_{ij}(X_i, X_j) + \lambda_{ij}(X_i) + \lambda_{ji}(X_j)$ define a *re-parameterization* of the original distribution, i.e., they change the individual functions without modifying the complete function $C(\mathbf{X})$. Depending on the literature, the $\lambda_{ij}$ are interpreted as "cost-shifting" operations that transfer cost from one function to another while preserving the overall cost function (Schiex, 2000; Rollon & Larrosa, 2006), or as Lagrange multipliers enforcing consistency among the copies of $X_i$ (Yedidia *et al.*, 2004; Wainwright *et al.*, 2005). In the former interpretation, the updates are called "soft arc-consistency" due to their similarity to arc-consistency for constraint satisfaction (Cooper & Schiex, 2004). Under the latter view, the bound corresponds to a *dual decomposition* solver for a linear programming (LP) relaxation of the original problem (Komodakis & Paragios, 2008; Sontag *et al.*, 2010).

The main distinguishing feature among such dual decomposition approaches is the way in which the bound is tightened by updating[2] the functions $\lambda$, generally either sub-gradient or gradient approaches (Komodakis & Paragios, 2008; Jojic *et al.*, 2010) or coordinate descent updates that can be interpreted as "message passing" (Globerson & Jaakkola, 2007; Sontag & Jaakkola, 2009). In practice, coordinate descent updates tend to tighten the bound more quickly, but can become caught in sub-optimal minima caused by

---

[2] We refer to these iterative bound improvement updates as "LP-tightening" updates, although technically we are tightening the decomposition bound (3) which is the dual of the LP. This is in contrast to literature that uses "tightening" to mean the inclusion of additional constraints (higher-order consistency), e.g., (Sontag *et al.*, 2008).

**Algorithm 2** LP-tightening

**Input:** Graphical model $\langle \mathbf{X}, \mathbf{D}, \mathbf{F}, \sum \rangle$, where $f_\alpha$ is a potential defined on variables $X_\alpha$.
**Output:** Upper bound on the optimum value
1: Iterate until convergence:
2: **for** any pair $\alpha, \beta$ with $X_{\alpha\beta} = X_\alpha \cap X_\beta \neq \emptyset$ **do**
3:   Compute max-marginals:
   $\gamma_\alpha(X_{\alpha\beta}) = \max_{X_\alpha \setminus X_{\alpha\beta}} f_\alpha(X_\alpha)$
   $\gamma_\beta(X_{\alpha\beta}) = \max_{X_\beta \setminus X_{\alpha\beta}} f_\beta(X_\beta)$
   Update parameterization:
   $f_\alpha(X_\alpha) \leftarrow f_\alpha(X_\alpha) + \frac{1}{2}\big(\gamma_\beta(X_{\alpha\beta}) - \gamma_\alpha(X_{\alpha\beta})\big)$
   $f_\beta(X_\beta) \leftarrow f_\beta(X_\beta) + \frac{1}{2}\big(\gamma_\alpha(X_{\alpha\beta}) - \gamma_\beta(X_{\alpha\beta})\big)$
4: **end for**

**Algorithm 3** Factor graph LP (FGLP)

**Input:** Graphical model $\langle \mathbf{X}, \mathbf{D}, \mathbf{F}, \sum \rangle$, where $f_\alpha$ is a potential defined on variables $X_\alpha$.
**Output:** Upper bound on the optimum value
1: Iterate until convergence:
2: **for** each variable $X_i$ **do**
3:   Get factors $F_i = \{\alpha : i \in \alpha\}$ with $X_i$ in their scope
4:   $\forall \alpha$, compute max-marginals:
   $\gamma_\alpha(X_i) = \max_{X_\alpha \setminus X_i} f_\alpha(X_\alpha)$
5:   $\forall \alpha$, update parameterization:
   $f_\alpha(X_\alpha) \leftarrow f_\alpha(X_\alpha) - \gamma_\alpha(X_i) + \frac{1}{|F_i|} \sum_{\beta \in F_i} \gamma_\beta(X_i)$
6: **end for**

the limited number of coordinate directions considered. Here we give an extremely simple update, most closely related to the "tree-block" coordinate descent updates derived in (Sontag & Jaakkola, 2009).

The algorithm is initialized with all $\lambda_{ij}(X_i) = 0$. Let us re-arrange the terms on right hand side of inequality 3, grouping together all functions that include a particular variable $X_i$ in their scope. Consider minimizing over a single pair $\lambda_{ij}(X_i), \lambda_{ik}(X_i)$; since all other terms are fixed, we minimize

$$\big[\max_{\mathbf{X}} f_{ij} + \lambda_{ij}\big] + \big[\max_{\mathbf{X}} f_{ik} + \lambda_{ik}\big]$$
$$= \max_{X_i} \big[\gamma_{ij}(X_i) + \lambda_{ij}(X_i)\big] + \max_{X_i} \big[\gamma_{ik}(X_i) + \lambda_{ik}(X_i)\big]$$
$$\geq \max_{X_i} \big[\gamma_{ij}(X_i) + \lambda_{ij}(X_i) + \gamma_{ik}(X_i) + \lambda_{ik}(X_i)\big]$$

where we have defined the "max-marginals" $\gamma_{ij}(X_i) = \max_{X_j} f_{ij}(X_i, X_j)$, and we require that $\lambda_{ij}(X_i) + \lambda_{ik}(X_i) = 0$ to preserve $\lambda \in \Lambda$. Many choices of $\lambda_{ij}$ achieve this minimum; a useful one is

$$\lambda_{ij} = \frac{1}{2}\big(\gamma_{ik}(X_i) - \gamma_{ij}(X_i)\big)$$

We can then set $f_{ij} \leftarrow f_{ij} + \lambda_{ij}$ and $\lambda_{ij} = 0$ (preserving $\lambda_{ij}(X_i) = 0$ for all $i,j$) and repeat, giving a simple coordinate descent update that can be interpreted as a max-marginal or "moment"-matching procedure on the functions $f_{ij}$. The update is easy to extend to higher-order functions $f_\alpha(X_\alpha)$; see Algorithm 2.

Another useful intuition comes from the alternative choice, $\lambda_{ij} = \gamma_{ik}(X_i) = \max_{X_k} f_{ik}(X_i, X_k)$. In this case, the contribution of edge $(ik)$'s term to the bound becomes zero. The "message" $\lambda_{ij}$ is equivalent to the dynamic program computed on the chain $j$–$i$–$k$. It is thus easy to see that dynamic programming (or variable elimination) on a tree can also be interpreted as coordinate descent on the same bound, using the same set of coordinates, and that this procedure will exactly optimize any tree-structured graph (Johnson et al., 2007; Yarkony et al., 2010).

A well-known LP-tightening algorithm is message-passing linear programming (MPLP) (Globerson & Jaakkola, 2007); its generalized version for higher-order cliques tightens the same bound as Algorithm 2 under descent updates along the same coordinates. The main distinction is that MPLP is formulated as sending messages between variable "beliefs" and the $f_{ij}$. Message-passing has the advantage that scheduled updates (e.g., Elidan et al., 2006; Sutton & McCallum, 2007) are easily applied. However, for cliques with large separator sets these messages can consume a noticable fraction of available memory. In contrast, re-parameterization updates like Algorithm 2 never store the $\lambda$, giving a memory advantage for large cliques. Although some scheduling approaches, such as those based on the decoding gap (Tarlow et al., 2011), are still applicable to pure reparameterization updates, our implementation used a fixed update ordering.

The original MPLP, soft-arc consistency, and many other LP algorithms operate directly on the original functions $f_\alpha$, updating coordinates that consist of single variable functions $\lambda(X_i)$ at each step; this corresponds to message passing on the "factor graph" representation of the model. Algorithm 3 (FGLP) gives a simple extension of the matching update that operates on this factor graph LP and tightens all $f_{ij}$ involved with some $X_i$ simultaneously. FGLP is similar to the "star-shaped" tree block coordinate descent derived in (Sontag & Jaakkola, 2009), but centered on separators (variables) rather than cliques (factors); in practice on our problems we found it faster than other update methods for this LP.

## 3  JOIN GRAPH LINEAR PROGRAMMING

From the perspective of Section 2.2, mini-bucket can be viewed within the LP-tightening framework. The mini-bucket procedure defines a join graph represented as a collection of maximal cliques and separators. The mini-bucket computations define a downward sweep of dynamic programming on the join graph with du-plicate variables; it is equivalent to running an LP-tightening procedure to convergence, *only* messages along edges of the mini-bucket spanning tree.

Given this view, it is straightforward to consider iterative updates on the same set of maximal cliques. We can construct a join graph using the mini-bucket relaxation, again assigning the original functions $f$ to their earliest clique and defining for any mini-bucket $q_i^k$ a function associated with its corresponding clique,

$$F_{q_i^k}(X_{q_i^k}) = \sum_{f_\alpha \in q_i^k} f_\alpha(X_\alpha) \qquad (4)$$

We then perform re-parameterization updates to the functions $F_q$ as in Algorithm 2. The resulting join-graph linear programming (JGLP) algorithm is shown in Algorithm 4.

Note that once JGLP converges, re-processing these new functions using mini-bucket elimination using the same value of $z$ will not change the choice of cliques (since no two cliques $q$, $q'$ could be joined without violating the clique size bound, $|X_q| \leq z + 1$). Additionally, the MBE pass will not change the bound value, since the MBE dynamic program can be viewed as a sequence of coordinate-descent updates along a subset of the edges, all of which must already be tight if the algorithm has converged. In the sequel, we will use this property of JGLP when developing heuristic evaluation functions for search; see Section 5.

**MBE vs. LP perspectives.** From the LP perspective, the MBE process can be thought of as a heuristic for selecting the cliques used to define the LP bound. This heuristic has the advantage of being very fast and controlled by a single integer value that is easy to search over, and it is easy to estimate the memory requirements of the approximation. In practice this results in a "top-down" construction, in which $z$ is set to the induced width and reduced until the computational resource constraints are met. Existing heuristics for selecting variable orderings with low induced width (Robertson & Seymour, 1983; Kask et al., 2011; Bodlaender, 2007; Gogate & Dechter, 2004) can be easily applied as well. In contrast, most existing generalized LP solvers work in a "bottom-up" fashion, running the LP to convergence on the original graph, then proposing slightly larger cliques (for example, from among fully connected triplets of variables (Sontag et al., 2008)) based on some greedy heuristic. However, for problems with small variables and reasonable resource constraints, it is easy to represent very large cliques ($z = 15$ to 25), in which case the top-down approach can be far more effective.

From the MBE perspective, the LP tightening updates provide an iterative version of mini-bucket that attempts to compensate for the variable copying relaxation. From this view it is most closely related to non-iterative cost-shifting procedures during the mini-bucket construction phase (Rollon & Larrosa, 2006). Another iterative "relax and compensate" approach was recently proposed in (Choi & Darwiche, 2009), but did not explicitly maintain a re-parameterization or use coordinate descent (thus could fail to converge).

**Algorithm 4** Algorithm JGLP
**Input:** Graphical model $\langle \mathbf{X}, \mathbf{D}, \mathbf{F}, \sum \rangle$; variable order $o = \{X_1, \ldots, X_n\}$; parameter $z$.
**Output:** Upper bound on the optimum value
1: Place each function $f_\alpha$ in its latest bucket in $o$
2: Build mini-bucket join graph (Algorithm 1)
3: Find the function of each mini-bucket as in (4)
4: **Iterate** to convergence / time-limit:
5: **for** all pairs $q^i$, $q^j$ connected by an edge **do**
6:    Find common variables, $S = var(q^i) \cap var(q^j)$
7:    Find the max-marginals of each mini-bucket $q^k$: $\gamma_k = \max_{var(q^k) \setminus S}(F_k)$, $k = i, j$
8:    Update functions in both mini-buckets
    $q^i$: $F_i \leftarrow F_i - \frac{1}{2}(\gamma_i - \gamma_j)$
    $q^j$: $F_j \leftarrow F_j + \frac{1}{2}(\gamma_i - \gamma_j)$
9: **end for**
10: **Return:** $\hat{C}^* = \sum_i \max_{X_i} F_i(X_i)$

## 4 MBE-MM

While the iterative nature of JGLP is appealing, it can have significant additional time and space overhead compared to MBE. We would thus also like to consider a single-pass MBE-like algorithm in which LP tightening is performed only within each bucket. We call the resulting algorithm *mini-bucket with max-marginal matching*, or MBE-MM.

MBE-MM proceeds by following the standard mini-bucket downward pass. However, when each mini-bucket $q_i^j \in Q_i$ is processed, eliminating variable $X_i$, we first perform an LP-tightening update to the mini-bucket functions $f_q$. For storage and computational efficiency reasons, we perform a single update on all buckets simultaneously, matching their max-marginals on their joint intersection. See Algorithm 5.

Viewing the matching update as a cost-shifting procedure, our MBE-MM algorithm is closely related to the work of (Rollon & Larrosa, 2006), who also proposed several dynamic cost-shifting updates for mini-bucket, with different update patterns. Their updates correspond to a "dynamic programming"-like heuristic that shifts all cost into a single function. In our case, we mimic the "balanced" cost shifting used by optimal iterative tightening but restrict our updates to a single bucket rather than to the whole global problem.

It is worth noting that, although any max-marginal

**Algorithm 5** Algorithm MBE-MM

**Input:** Graphical model $\langle \mathbf{X}, \mathbf{D}, \mathbf{F}, \sum \rangle$; variable order $o = \{X_1, \ldots, X_n\}$; parameter $z$.
**Output:** Upper bound on the optimum value
1: Place each function $f_\alpha$ in its latest bucket in $o$
2: **for** $i \leftarrow n$ down to 1 (processing bucket $\mathbf{B}_i$) **do**
3:     Partition functions in $\mathbf{B}_i$ into $Q_i = \{q_i^1, \ldots, q_i^p\}$, where each $q_i^k$ has no more than $z+1$ variables.
4:     Find the set of variables common to all the mini-buckets: $S_i = S_i^1 \cap \cdots \cap S_i^p$, where $S_i^k = var(q_i^k)$
5:     Find the function of each mini-bucket
$q_i^k$: $F_{ik} \leftarrow \prod_{f \in q_i^k} f$
6:     Find the max-marginals of each mini-bucket
$q_i^k$: $\gamma_{ik} = \max_{var(q_i^k) \setminus S_i}(F_{ik})$
7:     Update functions of each mini-bucket
$q_i^k$: $F_{ik} \leftarrow F_{ik} - \gamma_{ik} + \frac{1}{p} \sum_l \gamma_{il}$
8:     Generate messages $\lambda_i^k = \max_{X_i} F_{ik}$ and place each in the latest variable in $var(q_i^k)$'s bucket.
9: **end for**
10: **Return:** The buckets and cost bound from $\mathbf{B}_1$

matching step strictly tightens the fully decomposed bound (3) within each bucket, it does not necessarily tighten the overall solution found by MBE (which corresponds to fully optimizing the MBE subtree's edges); thus MBE-MM is not guaranteed to be tighter than ordinary MBE. However, it is reasonable to expect that the update will help, and in practice we find that the bounds are almost always significantly improved (see the experiments, Section 6).

Like "soft" arc-consistency, our algorithms are also related to known methods in constraint satisfaction. MBE($z$) and JGLP($z$), parameterized by a $z$-bound, are analogous to directional $z$-consistency and full $z$-consistency, respectively. Our algorithm MBE-MM represents an intermediate step between these two, and is analogous to an improvement of directional $i$-consistency with full iterative relational consistency schemes within each bucket (Dechter, 2003).

## 5 HEURISTICS FOR SEARCH

Mini-bucket is also a powerful mechanism for generating heuristics for informed search algorithms (Kask & Dechter, 2001). The intermediate functions $\lambda$ recorded by MBE (see Figure 1(b)) are used to express upper bounds on the best extension of any partial assignment, and so can be used as admissible heuristics guiding Best-First or Branch and Bound search.

Algorithms JGLP and FGLP do not produce heuristic functions directly; we obtain one by applying MBE to the modified (re-parameterized) functions output by the iterative algorithms. As noted in Section 3, for JGLP constructed with the same clique size bound $z$, this additional pass does not change the value of the bound. In contrast, applying MBE to the (much smaller) functions re-parameterized by FGLP forms new clusters and typically tightens the bound, yielding a hybrid heuristic generator "FGLP+MBE". Being a straightforward extension of "ordinary" MBE, the MBE-MM algorithm also directly yields a heuristic function suitable for informed search.

In our experiments (Section 6.2) we apply the output bounds as admissible heuristics for one of the most effective informed search approaches: the AND/OR Branch and Bound (AOBB) algorithm (Marinescu & Dechter, 2009a). This algorithm explores in a depth-first manner an AND/OR search space that is defined using a pseudo-tree arrangement of the problem's primal graph and takes advantage of the problem decomposition. The AND/OR search space is usually much smaller than the corresponding standard OR search space. AOBB keeps track of the value of the best solution found so far (a lower bound on the optimal cost) and uses this value and the heuristic function to prune away portions of the search space that are guaranteed not to contain the optimal solution in a typical branch-and-bound manner. For details on AOBB guided by MBE see (Marinescu & Dechter, 2009a).

## 6 EMPIRICAL EVALUATION

We investigate the impact of single-pass and iterative LP-tightening in conjunction with the MBE scheme for the task of finding the most probable explanation (MPE) over Bayesian networks. Specifically, we evaluate the performance of MBE-MM, Factor Graph LP-Tightening (denoted FGLP) applied to the original functions, and Join Graph LP-Tightening (JGLP) which is applied over the mini-bucket-based join graph. We compare these three algorithms against each other and against "pure" MBE, both as stand-alone bounding algorithms (Section 6.1) and as generator of heuristic evaluation functions that guides search algorithms such as Branch and Bound (Section 6.2).

Our benchmark problems include three sets of instances from genetic linkage analysis networks (Fishelson & Geiger, 2002) (denoted `pedigrees`, `type4b` and `LargeFam`) and grid networks from the UAI 2008 competition (Darwiche *et al.*, 2008). In total we evaluated 10 `pedigrees`, 10 `type4` instances, 40 `LargeFam` instances and 32 `grid` networks. The algorithms were implemented in C++ (64-bit).

### 6.1 LP-TIGHTENING ALGORITHMS AS BOUNDING SCHEMES

We compare the upper bounds' accuracy obtained by the non-iterative MBE and MBE-MM schemes and against the iterative FGLP and JGLP schemes. The iterative schemes ran for 5, 300 and 3600 seconds.

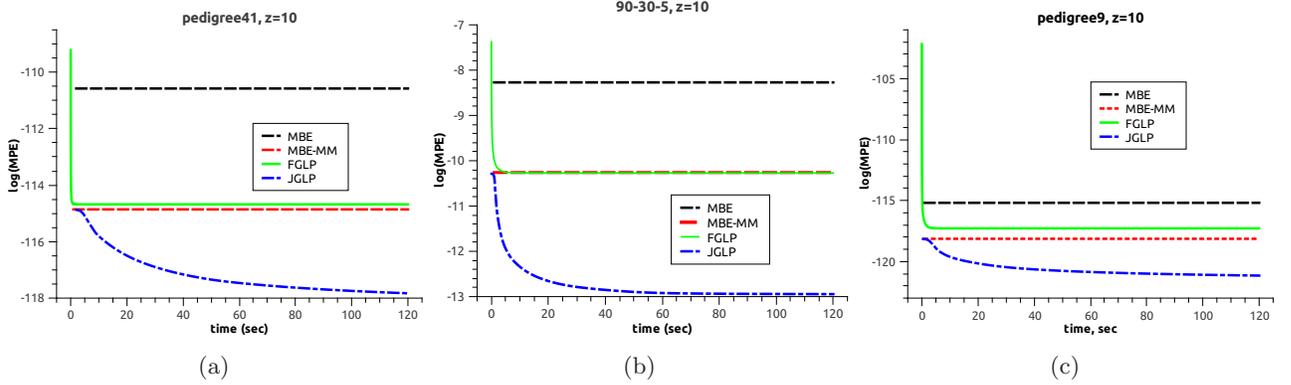

Figure 2: Upper bound on optimum (log scale) as a function of time (sec). Non-iterative bounds MBE, MBE-MM are shown for comparison.

Table 1: Upper bound (log scale) and runtime (# seconds) for a typical set of instances, $z = 10$ and $z = 20$. FGLP is not affected by $z$. Lower values are better. OOM shows that the algorithm ran out of memory (4 Gb). We report the number of variables $n$, largest domain size $k$, and the induced width $w$ along the ordering used.

| instance name | n | k | w | z | MBE UB/time | MBE-MM UB/time | FGLP time cut-offs | | | JGLP time cut-offs | | |
|---|---|---|---|---|---|---|---|---|---|---|---|---|
| | | | | | | | 5 UB | 300 UB | 3600 UB | 5 UB | 300 UB | 3600 UB |
| 75-25-5 | 625 | 2 | 34 | 10 | -15.4553/1 | -18.4089/1 | -16.6853 | -16.6854 | -16.6854 | -20.0289 | -20.8364 | -20.8364 |
| | | | | 20 | -17.4417/4 | -20.0576/4 | | | | -20.0576 | -20.1278 | -20.7067 |
| 90-30-5 | 900 | 2 | 42 | 10 | -8.2481/1 | -10.2597/1 | -10.2450 | -10.2705 | -10.2705 | -11.8469 | -12.9594 | -13.015 |
| | | | | 20 | -9.7424/7 | -11.6004/7 | | | | -11.6004 | -11.6942 | -12.5259 |
| 90-34-5 | 1156 | 2 | 48 | 10 | -8.42007/1 | -10.3708/1 | -9.65003 | -9.69458 | -9.69458 | -12.3469 | -13.2262 | -13.2883 |
| | | | | 20 | -9.58332/8 | -12.3670/9 | | | | -12.3670 | -12.5621 | -13.1538 |
| 90-42-5 | 1764 | 2 | 60 | 10 | -12.7401/1 | -15.9680/1 | -15.2480 | -15.3653 | -15.3653 | -18.4100 | -20.7714 | -20.8136 |
| | | | | 20 | -14.6136/13 | -18.5487/14 | | | | -18.5487 | -18.7679 | -19.9705 |
| largeFam4_11_51 | 1002 | 4 | 40 | 10 | -201.136/1 | -211.656/1 | -201.582 | -201.673 | -201.673 | -211.671 | -216.500 | -217.176 |
| | | | | 20 | OOM | OOM | | | | OOM | OOM | OOM |
| largeFam4_11_55 | 1114 | 4 | 38 | 10 | -229.43/1 | -242.489/1 | -226.075 | -226.328 | -226.328 | -242.657 | -249.551 | -250.453 |
| | | | | 20 | OOM | OOM | | | | OOM | OOM | OOM |
| largeFam4_12_51 | 1461 | 4 | 56 | 10 | -218.229/2 | -239.896/3 | -217.564 | -217.740 | -217.740 | -239.896 | -245.900 | -253.153 |
| | | | | 20 | OOM | OOM | | | | OOM | OOM | OOM |
| pedigree7 | 867 | 4 | 32 | 10 | -105.854/1 | -109.569/1 | -110.179 | -110.187 | -110.187 | -109.960 | -110.810 | -111.293 |
| | | | | 20 | -108.011/33 | -111.120/42 | | | | OOM | OOM | OOM |
| pedigree13 | 888 | 3 | 32 | 10 | -69.0973/1 | -70.0999/1 | -71.8561 | -71.8591 | -71.8591 | -70.4581 | -71.9869 | -72.0374 |
| | | | | 20 | -69.8890/8 | -71.1071/11 | | | | -71.1071 | -71.1071 | -71.3658 |
| pedigree31 | 1006 | 5 | 30 | 10 | -125.032/1 | -126.629/1 | -126.667 | -126.678 | -126.678 | -126.644 | -129.158 | -129.277 |
| | | | | 20 | OOM | OOM | | | | OOM | OOM | OOM |
| pedigree41 | 885 | 5 | 33 | 10 | -110.156/1 | -114.858/1 | -114.681 | -114.681 | -114.681 | -115.050 | -118.133 | -118.419 |
| | | | | 20 | -112.153/29 | -117.638/37 | | | | OOM | OOM | OOM |
| type4_120_17 | 4302 | 5 | 23 | 10 | -1128.22/1 | -1203.08/1 | -1049.34 | -1049.85 | -1049.86 | -1203.21 | -1221.21 | -1223.69 |
| | | | | 20 | -1235.94/18 | -1237.95/21 | | | | -1237.95 | OOM | OOM |
| type4_170_23 | 6933 | 5 | 21 | 10 | -1682.9/1 | -1747.18/1 | -1509.96 | -1511.61 | -1511.65 | -1747.22 | -1769.96 | -1772.16 |
| | | | | 20 | -1783.18/7 | -1783.76/7 | | | | -1783.76 | -1783.76 | -1783.76 |

In Table 1 we present a subset of the results obtained from all 4 instance sets for the $z$-bound values of $z = 10$ and $z = 20$. Results for $z = 15$ are similar and are omitted for lack of space. Note, that the value of $z$ does not influence the results of FGLP, that runs on the original functions. For every problem instance described via its parameters (i.e., number of variables $n$, largest domain size $k$, and induced width $w$), for each algorithm we report the upper bound obtained and the CPU time (or time-bound) in seconds.

We observe clearly that for all instances MBE-MM is superior to pure MBE, since it produces considerably more accurate bounds in a comparable time. For most instances the MBE-MM bounds are also better than pure FGLP (at the comparable point in time), especially for $z = 20$, but only if the memory required by MBE-MM is not too high. As example behavior see instances 75-25-5 and largeFam4_11_51.

Figure 2 illustrates the anytime performance of the iterative schemes FGLP and JGLP. (Note that the single-pass algorithms MBE and MBE-MM do not change as a function of time.) As we expect, FGLP improves the bound rapidly, but converges to a suboptimal solution, while JGLP improves at a slower pace but eventually produces the most accurate results.

From both the table and the figure we see that given enough time and memory, JGLP produces the most accurate bounds eventually. However, when time and memory are bounded, MBE-MM can present a cost-effective hybrid of bounded inference (as indicated by the cluster size) and bounded LP-tightening. Note that the memory requirement of MBE-MM is slightly less than the comparable JGLP, since some memory can be freed by the single-pass algorithm; this behavior is evidenced in pedigree7 and pedigree41 for $z = 20$.

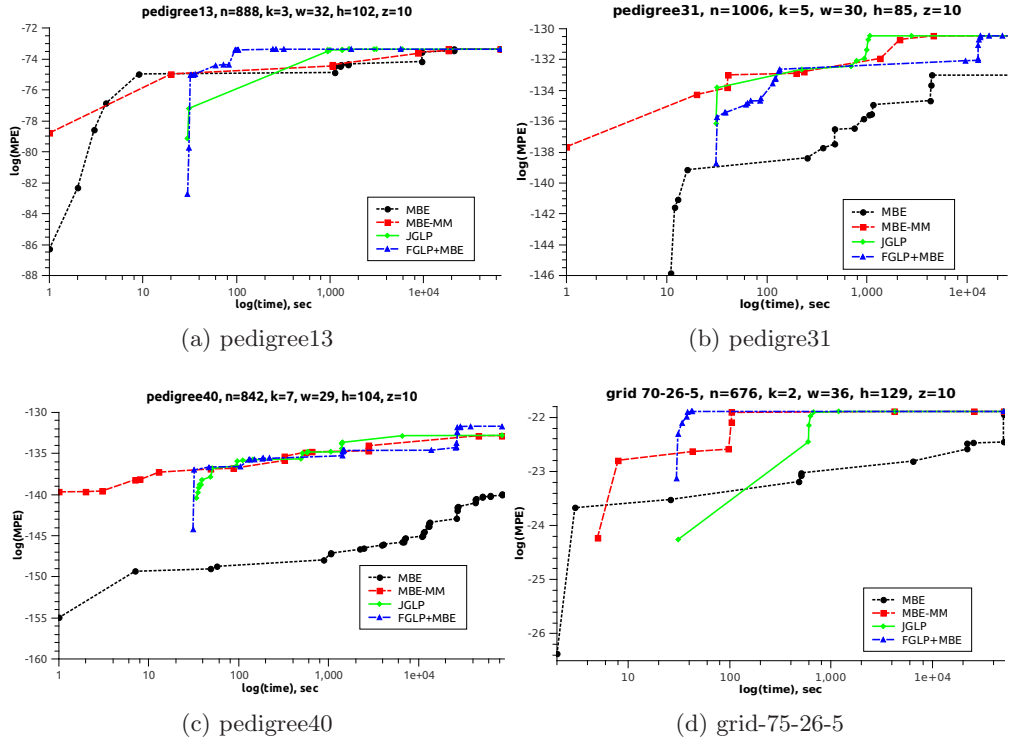

Figure 3: Lower bounds by the AOBB as a function of time (sec) for $z=10$, memory limit 3 Gb, timelimit 24h.

### 6.2 LP-TIGHTENING ALGORITHMS AS SEARCH GUIDING HEURISTICS

We also evaluated the impact of each of the bounding schemes as a generator of heuristic evaluation functions for the AOBB algorithm, as described in Section 5. We tested four schemes: AOBB guided by heuristics generated by pure MBE (denoted "AOBB-MBE"), AOBB guided by MBE and max-marginal matching heuristics ("AOBB-MBE-MM"), AOBB whose heuristics are generated from FGLP followed by MBE ("AOBB-FGLP+MBE") and AOBB guided by JGLP-produced heuristics ("AOBB-JGLP). All the heuristic functions were generated in a pre-processing phase, prior to search. The iterative FGLP and JGLP algorithms were run for 30 seconds each. The total time bound for AOBB with each of the heuristics was set to 24 hours (including the pre-processing), memory limit was 3 Gb, and the mini-bucket $z$-bound parameters for generating the heuristics was set to $z \in \{10, 15, 20\}$.

In Table 2 we show some of the results. For space reasons and clarity we pick a representative set from the full 92 instances. The table reports the total runtime in seconds and the number of nodes expanded by each of the AOBB schemes.

We see that, as expected, the heuristic generated by MBE-MM are more cost-effective compared with "pure" MBE, both in runtime and nodes expanded. We see that the two iterative schemes are quite powerful as heuristic generators. For most instances FGLP+MBE presents a good balance[3]. But on some hard instances and as long as memory is available JGLP is the overall best-performing scheme.

We also observe the impact of the $z$ bounds. As expected, with larger $z$ we obtain more accurate heuristics, but this increases the memory requirements (see, for example, AOBB-JGLP on pedigree13).

Figure 3 shows the anytime behaviour of AOBB with each heuristic, plotting the solution value found over time (both on log scale) on four typical instances. Higher values are better. As expected, AOBB-MBE-MM has a more precise heuristic and consistently outputs better solutions faster compared with AOBB-MBE. Algorithms AOBB-FGLP-MBE and AOBB-JGLP perform very well as anytime schemes. We see that they output the first solution later than the other two schemes due to initial 30 seconds pre-processing step. However, if desired, a shorter time bound for computing heuristic can be used.

---
[3]We did not test a "pure" FGLP heuristic, since its performance should be inferior to FGLP-MBE (FGLP followed by pure MBE).

Table 2: Search time (seconds) / # nodes expanded for selected instances. FGLP and JGLP ran for 30 seconds. "OOM" indicates that search ran out of memory (3Gb) and "— / —" that it ran out of time (24h). In **bold** we highlight the best runtime for each instance, *italics* indicate the smallest search space explored.

| Instances (n,k,w,h) | AOBB-MBE(z)<br>AOBB-MBE-MM(z)<br>AOBB-FGLP+MBE(z)<br>AOBB-JGLP(z)<br>z-bound=10<br>time / # nodes | AOBB-MBE(z)<br>AOBB-MBE-MM(z)<br>AOBB-FGLP+MBE(z)<br>AOBB-JGLP(z)<br>z-bound=15<br>time / # nodes | AOBB-MBE(z)<br>AOBB-MBE-MM(z)<br>AOBB-FGLP+MBE(z)<br>AOBB-JGLP(z)<br>z-bound=20<br>time / # nodes |
|---|---|---|---|
| **pedigree instances** | | | |
| pedigree7<br>(867, 4, 32, 90) | — / —<br>2171 / 348425451<br>805 / 140665826<br>530 / 80597149 | 9404 / 1876188145<br>428 / 78953096<br>**227** / *36619862*<br>286 / 38350755 | OOM<br>OOM<br>OOM<br>OOM |
| pedigree13<br>(888, 3, 32, 102) | — / —<br>66156 / 11726505961<br>5658 / 905160506<br>5911 / 1015334227 | 22799 / 5614980160<br>8150 / 1441111422<br>926 / 182970673<br>1939 / 366168237 | 7229 / 1522450313<br>704 / 164319080<br>**357** / *73658489*<br>OOM |
| pedigree31<br>(1006, 5, 30, 85) | — / —<br>61382 / 10617627744<br>24896 / 3695993630<br>2775 / 497649324 | — / —<br>3856 / 750931932<br>**1033** / *188749113*<br>2337 / 435238548 | OOM<br>OOM<br>OOM<br>OOM |
| **type4 linkage instances** | | | |
| type4b_120_17<br>(4072, 5, 24, 319) | — / —<br>— / —<br>— / —<br>— / — | — / —<br>— / —<br>— / —<br>— / — | — / —<br>**33** / *720778*<br>71 / 1168656<br>OOM |
| **LargeFam linkage instances** | | | |
| largeFam3_11_53<br>(1094, 3, 39, 71) | — / —<br>— / —<br>— / —<br>— / — | — / —<br>— / —<br>44663 / 8080262337<br>**10292** / *1878168857* | OOM<br>OOM<br>OOM<br>OOM |
| largeFam3_11_59<br>(1119, 3, 33, 73) | — / —<br>— / —<br>— / —<br>— / — | — / —<br>— / —<br>59012 / 8098379409<br>**22538** / *3025470612* | OOM<br>OOM<br>OOM<br>OOM |
| **binary grid instances** | | | |
| 90-30-5<br>(900, 2, 42, 151) | — / —<br>8601 / 1790747055<br>5928 / 1084067942<br>350 / 62930133 | 54415 / 10603123693<br>423 / 97620783<br>337 / 67303699<br>31 / 28688 | 5853 / 1299094138<br>**12** / 1125656<br>47 / 2101919<br>48 / *7493* |
| 90-42-5<br>(1764, 2, 60, 229) | — / —<br>— / —<br>— / —<br>**40** / 1411953 | — / —<br>62051 / 8399774202<br>17628 / 2349582057<br>134 / *13038792* | — / —<br>2471 / 340122171<br>651 / 93715978<br>OOM |
| 90-50-5<br>(2500, 2, 74, 312) | — / —<br>— / —<br>— / —<br>— / — | — / —<br>— / —<br>— / —<br>**48781** / *4187198638* | — / —<br>— / —<br>— / —<br>OOM |

In summary, based on our empirical evaluation we can conclude that some level of LP-tightening can significantly improve the power of the MBE heuristics, yielding an improved search, sometimes by orders of magnitudes. The question of instance-based balance, namely tailoring the right level of $z$-bound and LP-tightening to the problem instance is clearly a central issue and a direction of future research. Overall, in this study we observed that MBE-MM always improves upon MBE, using comparable time and memory, while FGLP quickly converges and is less memory-consuming than the other schemes. On the other hand, given sufficient time and memory JGLP produces the tightest bound.

## 7 CONCLUSION

The paper presents the first systematic combination of iterative cost-shifting updates with elimination-order based clustering algorithms and provides extensive empirical evaluation demonstrating its effectiveness. Specifically, we present Join Graph Linear Programming, a new bounding scheme for optimization tasks in graphical models that combines MBE bounds with LP-based cost-shifting or soft arc-consistency. Empirically, larger clusters improved the iterative updates' performance in all instances. We also demonstrated that JGLP can often find better bounds faster than LP-tightening on the original model, even for relatively small $z$. Most importantly, we showed the algorithms' ability to improve informed search algorithms; without requiring significantly more computational power (for fixed $z$) than classical MBE they can drastically reduce the search space. Notably, the algorithm that used as a heuristic generator all three cost-shifting schemes in a sequence (FGLP+JGLP+MBE-MM) won the first place in all optimization categories in this year's Probabilistic Inference Challenge.

**Acknowledgements.** Work supported in part by NSF Grant IIS-1065618 and NIH grant 5R01HG004175-03.

# References

Batra, D., Nowozin, S., & Kohli, P. 2011. Tighter relaxations for MAP-MRF inference: A local primal-dual gap based separation algorithm. *In: Conference on Uncertainty in Artificial Intelligence (AISTATS)*.

Bistarelli, S., Gennari, R., & Rossi, F. 2000. Constraint propagation for soft constraints: Generalization and termination conditions. *Principles and Practice of Constraint Programming–CP 2000*, 83–97.

Bodlaender, H. 2007. Treewidth: Structure and algorithms. *Structural Information and Communication Complexity*, 11–25.

Choi, Arthur, & Darwiche, Adnan. 2009. Approximating MAP by Compensating for Structural Relaxations. *Pages 351–359 of: Advances in Neural Information Processing Systems 22*.

Cooper, M., & Schiex, T. 2004. Arc consistency for soft constraints. *Artificial Intelligence*, **154**(1), 199–227.

Darwiche, A., Dechter, R., Choi, A., Gogate, V., & Otten, L. 2008. Results from the probablistic inference evaluation of UAI08, a web-report in http://graphmod.ics.uci.edu/uai08/Evaluation/Report. *In: UAI applications workshop*.

Dechter, R. 1999. Bucket elimination: A unifying framework for reasoning. *Artificial Intelligence*, **113**(1), 41–85.

Dechter, R. 2003. *Constraint processing*. Morgan Kaufmann.

Dechter, R., & Rish, I. 2003. Mini-buckets: A general scheme for bounded inference. *Journal of the ACM (JACM)*, **50**(2), 107–153.

Elidan, G., McGraw, I., & Koller, D. 2006. Residual Belief Propagation: Informed Scheduling for Asynchronous Message Passing. *In: Proc. UAI*.

Fishelson, M., & Geiger, D. 2002. Exact genetic linkage computations for general pedigrees. *Pages 189–198 of: International Conference on Intelligent Systems for Molecular Biology (ISMB)*.

Globerson, A., & Jaakkola, T. 2007. Fixing max-product: Convergent message passing algorithms for MAP LP-relaxations. *Advances in Neural Information Processing Systems*, **21**(1.6).

Gogate, V., & Dechter, R. 2004. A complete anytime algorithm for treewidth. *Pages 201–208 of: Proceedings of the 20th conference on Uncertainty in artificial intelligence*. AUAI Press.

Johnson, J., Malioutov, D., & Willsky, A. 2007. Lagrangian Relaxation for MAP Estimation in Graphical Models. *In: Allerton Conf. Comm. Control and Comput.*

Jojic, V., Gould, S., & Koller, D. 2010. Accelerated dual decomposition for MAP inference. *In: ICML*.

Kask, K., & Dechter, R. 2001. A general scheme for automatic generation of search heuristics from specification dependencies. *Artificial Intelligence*, **129**(1), 91–131.

Kask, K., Gelfand, A., Otten, L., & Dechter, R. 2011. Pushing the power of stochastic greedy ordering schemes for inference in graphical models. *AAAI 2011*.

Komodakis, N., & Paragios, N. 2008. Beyond loose LP-relaxations: Optimizing MRFs by repairing cycles. *Computer Vision–ECCV 2008*, 806–820.

Komodakis, N., Paragios, N., & Tziritas, G. 2007 (Oct.). MRF Optimization via Dual Decomposition: Message-Passing Revisited. *In: ICCV*.

Marinescu, Radu, & Dechter, Rina. 2009a. AND/OR Branch-and-Bound search for combinatorial optimization in graphical models. *Artificial Intelligence*, **173**(16-17), 1457–1491.

Marinescu, Radu, & Dechter, Rina. 2009b. Memory intensive AND/OR search for combinatorial optimization in graphical models. *Artificial Intelligence*, **173**(16-17), 1492–1524.

Mateescu, Robert, Kask, Kalev, Gogate, Vibhav, & Dechter, Rina. 2010. Join-Graph Propagation Algorithms. *Journal of Artificial Intelligence Research (JAIR)*, **37**, 279–328.

Pearl, J. 1988. *Probabilistic reasoning in intelligent systems: networks of plausible inference*. Morgan Kaufmann.

Robertson, N., & Seymour, P.D. 1983. Graph minors. I. Excluding a forest. *Journal of Combinatorial Theory, Series B*, **35**(1), 39–61.

Rollon, E., & Larrosa, J. 2006. Mini-bucket Elimination with Bucket Propagation. *Principles and Practice of Constraint Programming-CP 2006*, 484–498.

Schiex, T. 2000. Arc consistency for soft constraints. *Principles and Practice of Constraint Programming (CP2000)*, 411–424.

Sontag, D., Meltzer, T., Globerson, A., Jaakkola, T., & Weiss, Y. 2008. Tightening LP relaxations for MAP using message passing. *UAI*.

Sontag, D., Globerson, A., & Jaakkola, T. 2010. Introduction to dual decomposition for inference. *Optimization for Machine Learning*.

Sontag, David, & Jaakkola, Tommi. 2009. Tree Block Coordinate Descent for MAP in Graphical Models. *Pages 544–551 of: AI & Statistics*. JMLR: W&CP 5.

Sutton, C., & McCallum, A. 2007. Improved Dynamic Schedules for Belief Propagation. *Pages 376–383 of: Proc. UAI*.

Tarlow, Daniel, Batra, Dhruv, Kohli, Pushmeet, & Kolmogorov, Vladimir. 2011 (June). Dynamic Tree Block Coordinate Ascent. *Pages 113–120 of: ICML*.

Wainwright, M.J., Jaakkola, T.S., & Willsky, A.S. 2005. MAP estimation via agreement on trees: message-passing and linear programming. *Information Theory, IEEE Transactions on*, **51**(11), 3697–3717.

Yarkony, J., Fowlkes, C., & Ihler, A. 2010. Covering Trees and Lower Bounds on Quadratic Assignment. *In: CVPR*.

Yedidia, J. S., Freeman, W. T., & Weiss, Y. 2004 (May). *Constructing Free Energy Approximations and Generalized Belief Propagation Algorithms*. Tech. rept. 2004-040. MERL.